\documentclass[11pt, a4paper, logo, copyright, nonumbering]{xiaomi}

\usepackage[authoryear, sort&compress, round]{natbib}
\usepackage{dblfloatfix}
\usepackage{ulem}
\usepackage{caption}
\usepackage{dramatist}
\usepackage{xspace}
\usepackage{pifont} 
\usepackage{multirow}
\usepackage{tcolorbox}
\usepackage{xltabular}
\usepackage{longtable}
\usepackage{hyperref}
\usepackage{cleveref}
\usepackage{wrapfig}
\usepackage{graphicx}

\usepackage{amsfonts}
\usepackage{amsmath}
\usepackage{amssymb}
\usepackage{booktabs}
\usepackage{lineno}
\usepackage{multirow}
\usepackage{adjustbox}

\usepackage[bottom]{footmisc}

\usepackage{CJKutf8}
\usepackage{setspace}
\usepackage{makecell}
\usepackage{graphicx}
\usepackage{subcaption}
\usepackage{multicol} 

\definecolor{xiaomiorange}{HTML}{FF6901}

\allowdisplaybreaks[4]

\begin{abstract}
Reinforcement Learning with Verifiable Rewards (RLVR) has become a standard paradigm for reasoning in Large Language Models. However, optimizing solely for final-answer correctness often drives models into aimless, verbose exploration, where they rely on exhaustive trial-and-error tactics rather than structured planning to reach solutions. While heuristic constraints like length penalties can reduce verbosity, they often truncate essential reasoning steps, creating a difficult trade-off between efficiency and verification. In this paper, we argue that discriminative capability is a prerequisite for efficient generation: by learning to distinguish valid solutions, a model can internalize a guidance signal that prunes the search space. We propose JudgeRLVR, a two-stage \emph{judge-then-generate} paradigm. In the first stage, we train the model to judge solution responses with verifiable answers. In the second stage, we fine-tune the \emph{same} model with vanilla generating RLVR initialized from the judge. Compared to Vanilla RLVR using the same math-domain training data, JudgeRLVR achieves a better quality--efficiency trade-off for Qwen3-30B-A3B: on \textbf{in-domain math}, it delivers about \textbf{+3.7 points} average accuracy gain with \textbf{-42\%} average generation length; on \textbf{out-of-domain} benchmarks, it delivers about \textbf{+4.5 points} average accuracy improvement, demonstrating enhanced generalization.

\end{abstract}

\begin{document}

{
    \bgroup
    \setlength{\parindent}{0pt}
    \vspace*{3pt} 
    \begin{adjustwidth}{0pt}{0pt}  
    \begin{center} 
    {\titlefont JudgeRLVR: Judge First, Generate Second for Efficient Reasoning \par}
    \vskip5pt
    {
	{\normalfont\sffamily\fontsize{11}{15}\selectfont Jiangshan Duo$^{\dagger\ddagger\star}$ ~~~~~Hanyu Li$^{\ddagger\S}$ ~~~~~Hailin Zhang $^{\ddagger}$ } ~~~~~
	{\normalfont\sffamily\fontsize{11}{15}\selectfont Yudong Wang$^{\dagger\ddagger}$ \\ Sujian Li$^{\dagger\diamond}$ ~~~~~Liang Zhao$^{\ddagger\diamond}$} \\
	\vskip10pt
    {\normalfont\sffamily\fontsize{11}{15}\selectfont $^{\dagger}$State Key Laboratory of Multimedia Information Processing, School of } \\
	{\normalfont\sffamily\fontsize{11}{15}\selectfont  Computer Science, Peking University} \\
    {\normalfont\sffamily\fontsize{11}{15}\selectfont $^{\S}$CFCS, School of Computer Science, Peking University} \\
    {\normalfont\sffamily\fontsize{11}{15}\selectfont $^{\ddagger}$LLM-Core Xiaomi} 
	
	\vskip10pt
    }
    \end{center}
    \end{adjustwidth}
    \egroup
    {\abscontent}
    \thispagestyle{firststyle} 
}

\renewcommand{\thefootnote}{\fnsymbol{footnote}}
\footnotetext[0]{$^{\star}$Contribution during internship at Xiaomi LLM-Core Team.}
\footnotetext[0]{$^{\diamond}$Co-corresponding authors.}
\renewcommand{\thefootnote}{\arabic{footnote}}

\section{Introduction}
Reinforcement Learning with Verifiable Rewards (RLVR) has become a widely adopted method for improving large language models on complex reasoning tasks~\citep{Zhu2025, Shao2025}. By providing sparse yet automatically checkable supervision signals (e.g., whether the final answer is correct), RLVR enables models to explore strategies beyond supervised demonstrations and has achieved strong results on reasoning benchmarks like math and code.

\begin{figure}[!t]
  \centering
  \includegraphics[width=1.0\textwidth]{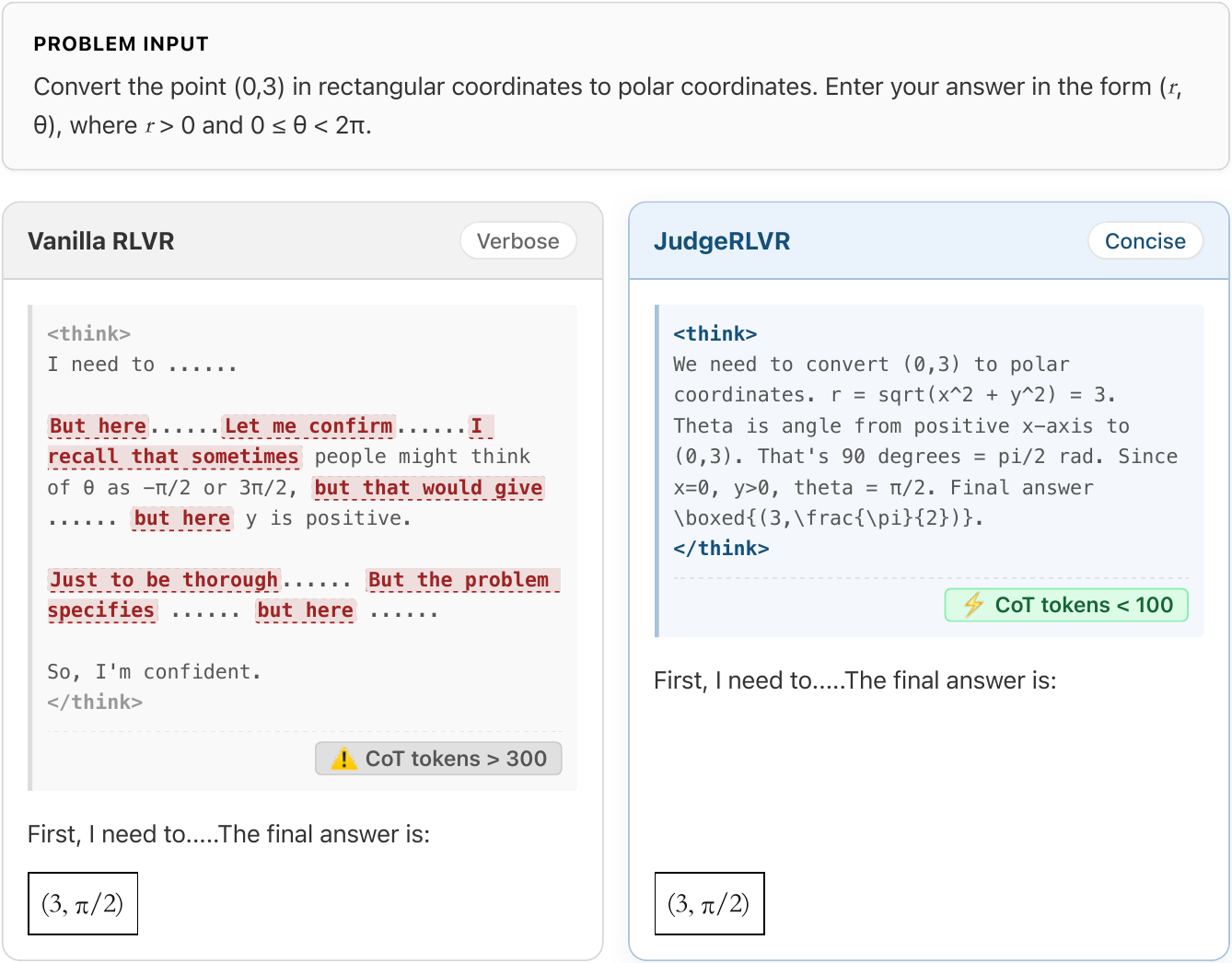}
  \caption{Qualitative comparison of reasoning patterns between Vanilla RLVR and JudgeRLVR. While outcome-based RL guarantees final answer correctness, it often induces verbose and unstructured exploration (left). JudgeRLVR implicitly regularizes the reasoning process, leading to a direct and coherent solution path (right).
  }
  \label{fig:JudgeRLVR_Example_Case}
\end{figure}

\begin{figure*}[t]
  \centering
  \includegraphics[width=1.0\linewidth]{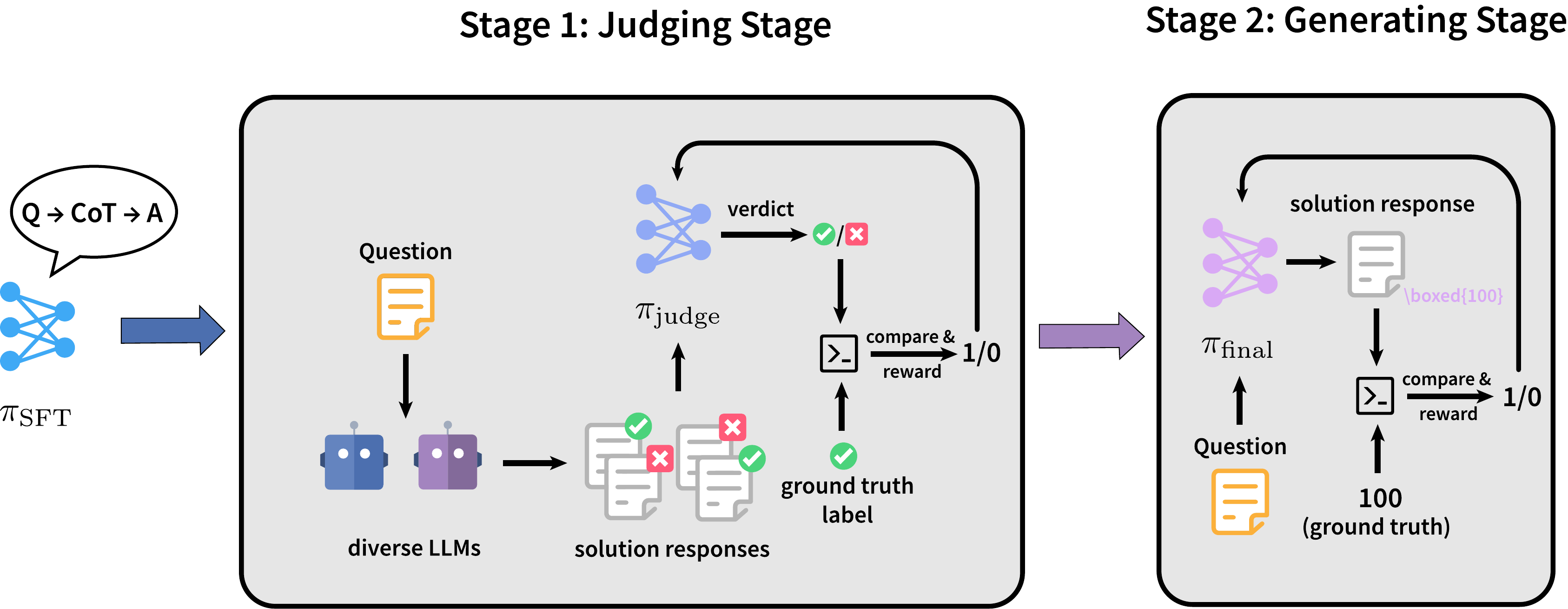}\hfill
  \caption{Pipeline for the two-stage training of JudgeRLVR}
  \label{fig:Pipeline}
\end{figure*}

However, RLVR does not merely optimize correctness; it also implicitly shapes a model's \emph{thinking pattern}. When trained primarily with a final-answer reward, many LLMs drift toward a generative search style: enumerating tentative branches, revising intermediate steps, and performing explicit self-correction~\citep{DeepSeekMath2024}. Figure~\ref{fig:JudgeRLVR_Example_Case} shows that the reasoning model's output can be divided into 2 parts: chain-of-thought (CoT) trajectories and solution responses. CoT trajectories usually include all the details of the model’s lengthy thought process, whereas solution responses tend to present a clear logical process along with the final answer. The resulting CoT trajectories are often long and exhibit frequent backtracking (e.g., heavy use of ``but'', ``however'', ``let's try again'')~\citep{Cai2025}. Long outputs are expensive, but a deeper issue is \emph{quality}: such trajectories are low in information density. 
Existing methods such as Kimi K1.5~\citep{Moonshot2025} and DAPO~\citep{yu2025dapo} introduce length penalties to reduce tokens and stabilize training, but often come with accuracy drops, making it difficult to achieve an ideal quality--efficiency trade-off.

This raises a more general question: can we explore a training paradigm that explicitly targets higher-quality thinking patterns, rather than hoping the model eventually ``stumbles upon'' them through repeated trial-and-error? Cognitive science suggests that expert reasoning is not defined by externalizing exhaustive search, but by \emph{early discrimination and pruning}: unpromising paths are filtered before they are expanded~\citep{Chi1981}. This aligns with common intuition: experts quickly spot low-value reasoning paths and focus on high-yield hypotheses, while novices resort to trial-and-error and frequent backtracking.
This intuition motivates the idea that, to achieve efficient reasoning, one should first develop the ability to judge what constitutes direct and valid reasoning, and then learn how to reason for specific problems.

Based on the idea above, we propose \textbf{JudgeRLVR}, a two-stage \emph{judge-then-generate} paradigm. As shown in Figure~\ref{fig:Pipeline}, in judging stage, we train the model as a judge: given candidate solution responses generated by diverse models, it learns to classify them as correct or incorrect. In generating stage, we initialize the model from the judging stage and apply \emph{Vanilla RLVR} fine-tuning on the same model. 
Under this training paradigm, the model learns to implicitly prune low-quality branches prior to generating extensive textual search, thereby fostering more direct and reliable reasoning patterns.

It is worth noting that, we do not impose any explicit length or quality penalties on the CoT trajectories; thus, any reduction in verbosity or refinement in reasoning style can be attributed  to the discriminative priors transferred  from the judging stage. 

With our training paradigm, a model can  achieve an optimal balance between performance and efficiency.
Compared to Vanilla RLVR using the same training data, JudgeRLVR achieves a quality--efficiency trade-off on Qwen3-30B-A3B~\citep{qwen3technicalreport}: on \textbf{in-domain math}, it achieves about \textbf{+3.7 points} average accuracy improvement while reducing average generation length by about \textbf{42\%}; on \textbf{out-of-domain} benchmarks, it delivers about \textbf{+4.5 points} average accuracy improvement.
From extensive experiments, we also analyze the specific impact of this training paradigm on thinking patterns. We observe a shift in the linguistic style of the generated text, with notably fewer explicit backtracking cues such as ``but'' during reasoning. This suggests that JudgeRLVR moves  from an externalized trial-and-error process to an internal decision-making process, thereby producing higher-quality reasoning patterns.

In summary, our main contributions are: (i) we propose \textbf{JudgeRLVR}, a new training paradigm for inducing higher-quality thinking patterns via \emph{judge-then-generate}; (ii) JudgeRLVR demonstrates a better quality--efficiency trade-off than Vanilla RLVR on both in-domain and out-of-domain tasks; 
(iii) we provide interpretable linguistic evidence which is consistent with reduced explicit backtracking.

\section{Related Work}
\subsection{RLVR and Verifiable-Reward Training for Reasoning}
Reinforcement Learning with Verifiable Rewards (RLVR) is a leading approach for improving LLM reasoning using automatic, task-specific rewards, often outperforming supervised fine-tuning on math and logic tasks~\citep{Lambert2024, Guo2025a, Xu2025}. However, RLVR has notable limitations: it can yield gains even under spurious rewards~\citep{Shao2025}, and it mainly reweights existing high-reward reasoning paths rather than inducing new reasoning behaviors~\citep{Yue2025}. This is consistent with concerns that RLVR may implicitly encourage suboptimal thinking by optimizing final correctness instead of process quality~\citep{DataLearner2025}. Most variants emphasize reward design or optimization, leaving the quality--efficiency trade-off underexplored.

\subsection{Mathematical Reasoning in Large Language Models}
Mathematical reasoning is a core benchmark for LLMs, where recent progress relies heavily on RLVR and targeted training~\citep{DeepSeekMath2024, Shen2025}. Data diversification via paraphrasing can improve generalization~\citep{Shen2025}, while extended RLVR training improves accuracy but tends to produce verbose reasoning traces~\citep{DeepSeekMath2024}. Synthetic data scaling further boosts capability yet does not eliminate redundancy~\citep{Zhou2025}, and scaling studies still report low information density in reasoning outputs~\citep{Sun2024}. Overall, improving both correctness and efficiency remains challenging.

\subsection{Learning to Judge: Evaluators, Verifiers, and Discriminative Supervision}
Training evaluators and using discriminative supervision can steer LLM outputs by separating good from bad generations~\citep{Guo2025b, Tang2025}. Discriminative fine-tuning can suppress negative outputs without preference data~\citep{Guo2025b}, and stronger evaluators improve reasoning-quality assessment via coherence-focused criteria~\citep{Tang2025}. Verifiers can detect reasoning errors with uncertainty signals~\citep{Han2024}, and systematic studies highlight the need for task-specific validation metrics~\citep{Shimabucoro2024}. Yet existing work rarely adopts a two-stage pipeline that first trains judgment and then uses it to guide generation, which we address in JudgeRLVR.

\section{Method}
\textbf{JudgeRLVR} divides the training of a single reasoning policy into two sequential stages with different roles but intended capability transfer: a \emph{judge} stage that trains discriminative error awareness, followed by a \emph{generating} stage that optimizes solution generation. Our core hypothesis is that discriminative competence is a prerequisite for efficient generation. Once the model internalizes what constitutes a valid CoT trajectory, it can down-weight erroneous branches early in generation without any explicit length penalty.

Figure~\ref{fig:Pipeline} illustrates the complete pipeline of our two-stage training.
\subsection{Notation and Task Definition}
The dataset consists of problems in a specific domain: $x \in \mathcal{X}$ with gold final answers $a^\star(x)$.

The \textbf{solution response} is denoted $z=(o_1, \dots, o_T)$, a token sequence containing a clear logical process and ending with a final answer string. The final answer $a(z)$ is extracted from the solution responses via a deterministic parser. We do not train the model to judge directly on full outputs that include complete CoT trajectories because such outputs are often extremely long and contain a large amount of distracting information that can interfere with the judgment. Instead, we expect the model to judge directly based on the clean solution response, so that it can identify errors with greater precision.

We define a binary correctness label $\ell \in \{0, 1\}$:
\begin{equation}
\ell(x, z) = \mathbb{I}\big(a(z) = a^\star(x)\big).
\end{equation}
The policy (language model) is denoted $\pi_\theta$.

\subsection{Stage~1: Judging Stage}
In judging stage, the model is trained as a judge. Given a problem $x$ and a candidate solution response $z$, the policy produces a critique/commentary $c$ which contains the CoT trajectory, and outputs a discrete verdict token $v \in \{0, 1\}$ (0 for incorrect and 1 for correct):
\begin{equation}
(c, v) \sim \pi_\theta(\cdot \mid x, z).
\end{equation}

\paragraph{Data construction.}
We construct a discriminative dataset $D_{\text{judge}}$ consisting of triplets $(x, z, \ell)$:
\begin{itemize}
  \item \textbf{Response generation (rollout).} For each problem $x$, we sample a set of candidate solution responses $\{z_1, \dots, z_n\}$ from several models, and obtain $\ell(x, z_i)$ by comparing $a(z_i)$ with $a^\star(x)$.
  \item \textbf{Hard negative mining.} We prioritize \emph{moderate-difficulty} problems whose empirical pass rates under rollouts is neither 0 nor 1, yielding more informative ``nearly-correct'' errors.
  \item \textbf{Class balancing.} For each $x$, we subsample to balance positives ($\ell=1$) and negatives ($\ell=0$), avoiding majority-class bias.
\end{itemize}

\paragraph{Reward.}
The reward is whether the final verdict $v_i$ matches the true label $\ell(x, z_i)$:
\begin{equation}
r_i = \mathbb{I}(v_i=\ell(x, z_i))
\end{equation}

The training objective encourages $v_i$ to match the true label $\ell(x, z_i)$. Meanwhile, the commentary $c$, as an explanatory prefix, is optimized by the same policy-gradient signal.

\subsection{Stage~2: Generating Stage}
In generating stage, the model is trained under Vanilla RLVR setting and outputs solutions of the problems. The policy is initialized from the judging stage weights.

Given a problem $x$, the policy generates a CoT trajectory and the solution response $z$ (including the final parsed answer $a(z)$):
\begin{equation}
z \sim \pi_\theta(\cdot \mid x).
\end{equation}
The reward remains the sparse binary correctness signal based only on the final answer:
\begin{equation}
r = \ell(x, z) = \mathbb{I}\big(a(z)=a^\star(x)\big).
\end{equation}

\subsection{Hypothesized Mechanism and Verification Overview}
We hypothesize that JudgeRLVR improves reasoning-pattern quality through the two stages in two corresponding ways: (i) \textbf{style transfer}: in judging stage, learning to judge induces a global shift in the model's reasoning language style; (ii) \textbf{reduced backtracking}: in generating stage, training activates the internalized efficient pattern, reducing explicit self-correction and backtracking in text. We design targeted experiments to test both points; definitions and results are provided in Section~\ref{sec:experiments}.

\section{Experimental Setup}
\label{sec:experiments}

\subsection{Model Configuration and Training Data}
We use Qwen3-30B-A3B~\citep{qwen3technicalreport}, a mixture-of-experts (MoE) architecture. To ensure a basic level of reasoning capability, we first conduct supervised fine-tuning (SFT) on curated open-source CoT datasets, obtaining Qwen3-30B-A3B-SFT with improved reasoning and instruction-following abilities.

For RLVR, we collect and filter open-source resources to build a dataset of 113k math problems with gold answers. To construct specialized training data for the judging stage, we adopt a strict sampling strategy:
\begin{itemize}
  \item \textbf{Rollout generation.} For each problem, we use MiMo-7B RL~\citep{coreteam2025mimounlockingreasoningpotential} and the target model Qwen3-30B-A3B-SFT to each generate 8 reasoning traces and answers, yielding 16 distinct reasoning paths.
  \item \textbf{Balanced sampling.} We filter outputs so that each problem retains an equal number of correct and incorrect solutions, avoiding bias from class imbalance.
\end{itemize}

Our training algorithm is DAPO~\citep{yu2025dapo}, a GRPO-family policy gradient method. 
We implement dynamic sampling to filter out samples with pass rates $0$ or $1$ during rollout. The batch size is set to 256 and the rollout size is $n=8$. We use AdamW optimizer~\citep{loshchilovDecoupledWeightDecay2018} for training. The learning rate is set to $3\times 10^{-6}$ with a $10$-step warmup. We use a training temperature of $1.0$ and a testing temperature of $0.6$, with top-$p$ set to $1.0$ and max tokens set to 65536 for both training and testing.

\subsection{Evaluation Benchmarks}
We evaluate JudgeRLVR on in-domain math reasoning: AIME24~\citep{AIME24}, AIME25~\citep{AIME25}, MATH500~\citep{Hendrycks2021MATH}, HMMT\_feb\_2025~\citep{HMMT2025}, BeyondAIME~\citep{BeyondAIME2025} and on diverse out-of-domain benchmarks: GPQA Diamond (Science Reasoning)~\citep{Rein2023}, IFEval (Instruction Following)~\citep{Zhou2023}, LiveCodeBenchv6 (Coding)~\citep{Jain2024}, MMLU-Redux (General Knowledge)~\citep{Gema2025}, ZebraLogic (Logical Reasoning)~\citep{Lin2025}, covering a wide range of general capabilities. The detailed setups of these benchmarks are in appendix~\ref{sec:Evaluate Details}.

\subsection{Main Experiments}
Our main experiments are designed to answer a single overarching question: does the proposed judge-then-generate paradigm improve the quality--efficiency trade-off of RLVR beyond Vanilla RLVR? To make this comparison meaningful, we evaluate three training settings that isolate the effect of the staged paradigm from other factors.

\textbf{Base SFT}: the static baseline model Qwen3-30B-A3B-SFT. Base SFT serves as a reference point for the model's reasoning ability before RL and its default generation style, allowing us to measure how much of the downstream gains come specifically from RLVR.

\textbf{Vanilla RLVR}: Vanilla generating RLVR optimizing only for final-answer correctness. Vanilla RLVR serves as a strong baseline to demonstrate the advantages of our method. The total training is 250 steps.

\textbf{JudgeRLVR}: our judge-then-generate method. The judging stage is 145 steps and the generating stage is 105 steps.

Vanilla RLVR and JudgeRLVR use the same training and evaluation hyperparameters and the same total training steps. We report both accuracy and generation length because our goal is not only to increase correctness, but also to reduce unnecessary computation and improve information density in the produced reasoning.

The generating and rollout prompt, and the judging prompt are in appendix~\ref{sec:prompt templates}.

\subsection{Ablations}
We conduct targeted ablations to validate that JudgeRLVR's gains come from the divided stages itself rather than from simply adding more training signals.

\textbf{Judge Only}: only judging stage training. Judge Only setting evaluates whether judge training alone can already improve generation quality, or whether generating stage is necessary to translate judging competence into improved generation performance. This ablation directly tests our core assumption that judging is a prerequisite but not a substitute for generating optimization.

\textbf{Mixed Strategy}: judging stage is replaced by parallel training on mixed Judge/Generative data (ratio 1:1), with each data type using its corresponding reward; generating stage is then run as usual. Mixed Strategy setting probes whether the ordering and separation of roles matters: by interleaving judge-style and generation-style updates, we test whether simultaneously optimizing two behaviors interferes with learning a clean internal decision policy. In other words, this ablation examines whether the model benefits from first consolidating an error-aware judge prior before being asked to explore and exploit it during generation. 

Together, these ablations aim to distinguish (i) ``benefit from judging'' from (ii) ``benefit from judging first, then generating, '' clarifying which component is responsible for improved efficiency and generalization.

\subsection{Mechanism Verification}
To test style transfer and reduced backtracking, we design:

\textbf{Perplexity (PPL).} PPL is the exponential of a language model’s average negative log-likelihood over a sequence; it measures how ``surprised'' or ``confused'' the model is, and lower values indicate the model is more familiar with the sequence content. Changes in perplexity can reflect the extent to which a model is familiar with a particular linguistic style. Using the Base SFT model as a fixed evaluator, we compute changes in the PPL of outputs produced under different training methods; these changes then reflect stylistic shifts during training. During judging stage training of JudgeRLVR, we sample model outputs at each step (10 samples per step) for Vanilla RLVR and JudgeRLVR, and compute the average PPL of the Base SFT model on these outputs.

\textbf{Transition-word statistics.} Behaviors such as explicit backtracking and reflection in model outputs are typically manifested through the use of transition-words. Therefore, by counting the number and frequency of such discourse markers in outputs over the course of training, these statistics can indicate, to some extent, that backtracking is decreasing. During generating stage training of JudgeRLVR, we sample 1000 outputs per step and count occurrences and frequencies of contrast/backtracking markers:
  \begin{quote}\small
  \texttt{but, however, though, although, yet, still, nevertheless, nonetheless, instead, conversely, whereas, while, actually, wait}.
  \end{quote}

  The above experiments are used to demonstrate shifts in style and backtracking. Combined with improvements in model performance, they can provide a comprehensive indication that these shifts are moving in a favorable direction.

\begin{table*}[t]
\centering
\caption{Main results comparing Base SFT, Vanilla RLVR, and JudgeRLVR (Sequential). We report accuracy (Acc) and average generation length in tokens (Len). For JudgeRLVR, $\Delta$ is absolute change in accuracy (pp:percentage point) vs Vanilla RLVR (higher is better); for Len, $\Delta$ is relative change vs Vanilla RLVR (shorter is better).}
\label{tab:main_results_base_gen_seq}
\resizebox{1.0\textwidth}{!}{%
\begin{tabular}{l|rr|rr|rrrr}
\toprule
 & \multicolumn{2}{c|}{\textbf{Base SFT}} & \multicolumn{2}{c|}{\textbf{Vanilla RLVR}} & \multicolumn{4}{c}{\textbf{JudgeRLVR (Seq)}} \\
\textbf{Benchmark} &
\textbf{Acc}$\uparrow$ & \textbf{Len}$\downarrow$ &
\textbf{Acc}$\uparrow$ & \textbf{Len}$\downarrow$ &
\textbf{Acc}$\uparrow$ & \textbf{$\Delta$ (pp)} & \textbf{Len}$\downarrow$ & \textbf{$\Delta$ (\%)} \\
\midrule
\multicolumn{9}{l}{\textit{In-Domain (Math)}} \\
\textbf{AIME24}
& 85.7 & 20.8k
& 86.3 & 21.8k
& \textbf{89.0} & +2.7 & \textbf{12.9k} & -41 \\
\textbf{AIME25}
& 74.1 & 26.8k
& 77.0 & 27.2k
& \textbf{78.7} & +1.7 & \textbf{17.2k} & -37 \\
\textbf{MATH500}
& 97.4 & 6.7k
& \textbf{98.0} & 6.8k
& 97.2 & -0.8 & \textbf{2.0k} & -71 \\
\textbf{HMMT\_feb\_2025}
& 57.1 & 24.9k
& 60.8 & 26.9k
& \textbf{70.0} & +9.2 & \textbf{20.4k} & -24 \\
\textbf{BeyondAIME}
& 55.5 & 33.8k
& 58.3 & 34.3k
& \textbf{63.9} & +5.6 & \textbf{21.3k} & -38 \\
\textbf{Overall Avg}
& 74.0 & 22.6k
& 76.1 & 23.4k
& \textbf{79.8} & +3.7 & \textbf{14.8k} & -42 \\
\midrule
\multicolumn{9}{l}{\textit{Out-of-Domain}} \\
\textbf{GPQA Diamond}
& 63.0 & 12.0k
& 61.2 & 12.1k
& \textbf{66.4} & +5.2 & \textbf{11.2k} & -7.5 \\
\textbf{IFEval}
& 80.6 & \textbf{3.2k}
& 79.9 & 4.1k
& \textbf{86.4} & +6.5 & 4.6k & +12 \\
\textbf{LiveCodeBenchv6}
& 57.4 & 20.6k
& 58.2 & 22.8k
& \textbf{63.9} & +5.7 & \textbf{18.7k} & -18 \\
\textbf{MMLU Redux}
& 86.8 & 2.1k
& 85.4 & 2.1k
& \textbf{87.0} & +1.6 & \textbf{1.5k} & -28 \\
\textbf{ZebraLogic}
& 84.6 & \textbf{7.7k}
& 83.6 & 7.9k
& \textbf{86.8} &  +3.2 & 9.6k & +22 \\
\textbf{Overall Avg}
& 74.5 &  9.1k
& 73.7 &  9.8k
& \textbf{78.1} & +4.5 & \textbf{ 9.1k} & -3.9 \\
\bottomrule
\end{tabular}%
}
\end{table*}

\begin{table*}[t]
\centering
\caption{Main results comparing JudgeRLVR (Sequential), Judge Only, and Mixed Strategy. We report accuracy (Acc) and average generation length in tokens (Len). For Judge Only and Mixed Strategy, $\Delta$ is absolute change in accuracy (pp:percentage point) vs JudgeRLVR (higher is better); for Len, $\Delta$ is relative change vs JudgeRLVR (shorter is better).}
\label{tab:main_results_mixed_judgeonly_seq}
\resizebox{\textwidth}{!}{%
\begin{tabular}{l|rr|rrrr|rrrr}
\toprule
 & \multicolumn{2}{c|}{\textbf{JudgeRLVR (Seq)}} & \multicolumn{4}{c|}{\textbf{Judge Only}} & \multicolumn{4}{c}{\textbf{Mixed Strategy}} \\
\textbf{Benchmark} &
\textbf{Acc}$\uparrow$ & \textbf{Len}$\downarrow$ &
\textbf{Acc}$\uparrow$ & \textbf{$\Delta$ (pp)} & \textbf{Len}$\downarrow$ & \textbf{$\Delta$ (\%)} &
\textbf{Acc}$\uparrow$ & \textbf{$\Delta$ (pp)} & \textbf{Len}$\downarrow$ & \textbf{$\Delta$ (\%)} \\
\midrule
\multicolumn{11}{l}{\textit{In-Domain (Math)}} \\
\textbf{AIME24}
& \textbf{89.0} & \textbf{12.9k}
& 87.0 & -2.0 & 22.4k & +74
& 88.3 & -0.7 & 12.8k & -0.8 \\
\textbf{AIME25}
& 78.7 & \textbf{17.2k}
& 76.3 & -2.4 & 28.5k & +66
& \textbf{79.7} & +1.0 & 18.2k & +6.0 \\
\textbf{MATH500}
& 97.2 & \textbf{2.0k}
& 97.2 & +0.0 & 7.4k & +266
& \textbf{98.0} & +0.8 & 2.2k & +7.9 \\
\textbf{HMMT\_feb\_2025}
& \textbf{70.0} & 20.4k
& 61.5 & -8.5 & \textbf{17.9k} & -12
& 69.0 & -1.0 & 25.9k & +27 \\
\textbf{BeyondAIME}
& \textbf{63.9} & \textbf{21.3k}
& 57.3 & -6.6 & 35.8k & +68
& 63.4 & -0.5 & 22.9k & +7.3 \\
\textbf{Overall Avg}
& 79.8 & 14.8k
& 75.9 & -3.9 & 22.4k & +92
& 79.7 & -0.1 & 16.4k & +9.5 \\
\midrule
\multicolumn{11}{l}{\textit{Out-of-Domain}} \\
\textbf{GPQA Diamond}
& 66.4 & 11.2k
& \textbf{66.9} & +0.5 & \textbf{7.9k} & -30
& 63.7 & -2.7 & 16.1k & +44 \\
\textbf{IFEval}
& \textbf{86.4} & 4.6k
& 85.9 & -0.5 & \textbf{4.2k} & -9.0
& 80.4 & -6.0 & 7.9k & +72 \\
\textbf{LiveCodeBenchv6}
& \textbf{63.9} & \textbf{18.7k}
& 61.4 & -2.5 & 21.3k & +14
& 56.4 & -7.5 & 26.7k & +43 \\
\textbf{MMLU Redux}
& 87.0 & 1.5k
& \textbf{87.4} & +0.4 & \textbf{1.0k} & -31
& \textbf{87.1} & +0.1 & 3.3k & +119 \\
\textbf{ZebraLogic}
& 86.8 & 9.6k
& 87.3 & +0.5 & 8.0k & -17
& \textbf{87.4} & +0.6 & \textbf{7.9k} & -19 \\
\textbf{Overall Avg}
& 78.1 & 9.1k
& 77.8 & -0.3 & 8.5k & -15
& 75.0 & -3.1 & 12.4k & +52 \\
\bottomrule
\end{tabular}%
}
\end{table*}

\begin{figure}[t]
  \centering
  \includegraphics[width=0.75\columnwidth]{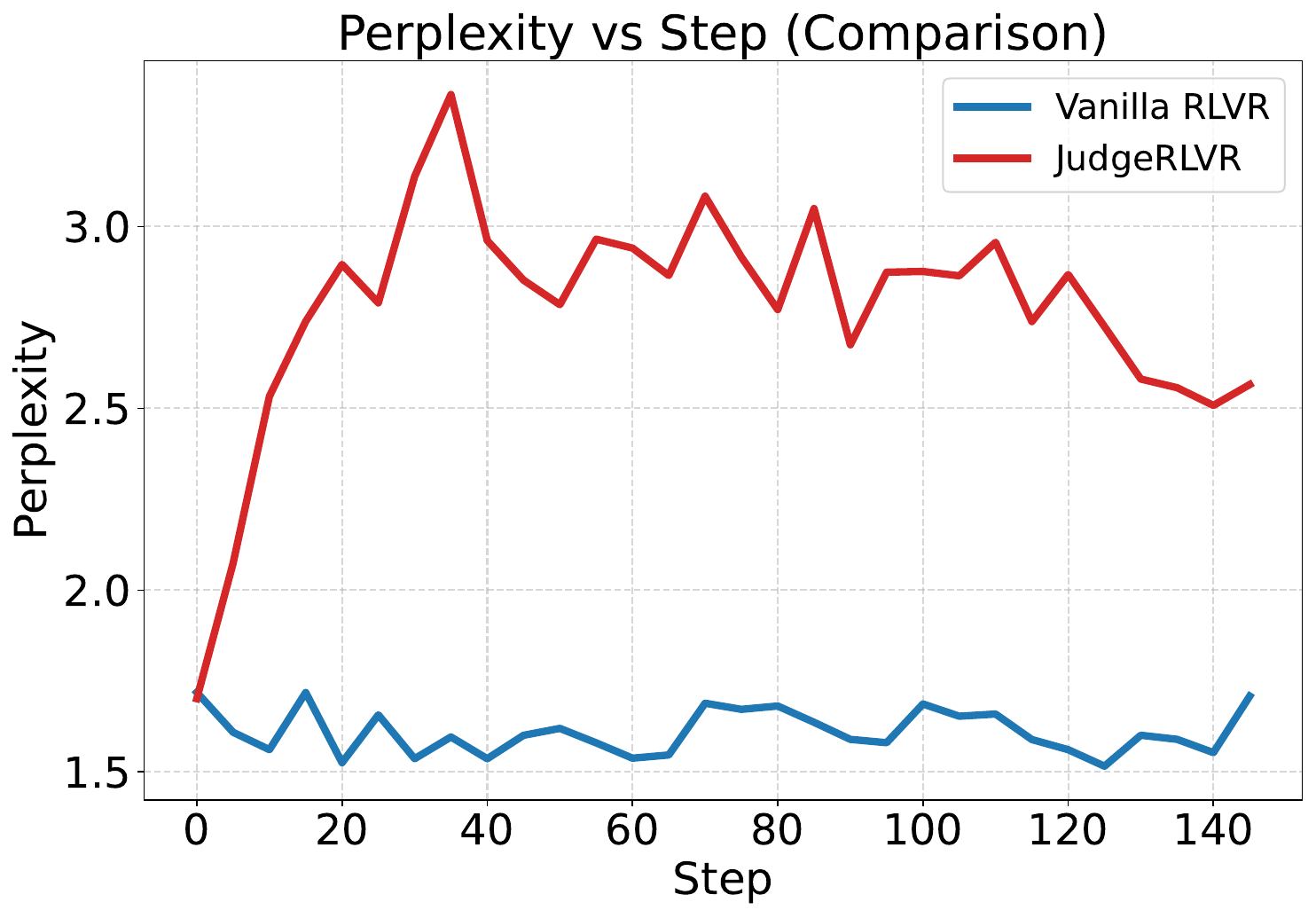}\hfill
  \caption{Base SFT perplexity (PPL) evaluated on sampled outputs along training steps. Vanilla RLVR stays stylistically close to Base SFT (flat PPL), while JudgeRLVR (judging stage) exhibits increasing PPL, indicating style transfer.}
  \label{fig:ppl_vs_step}
\end{figure}

\begin{figure*}[t]
  \includegraphics[width=0.5\linewidth]{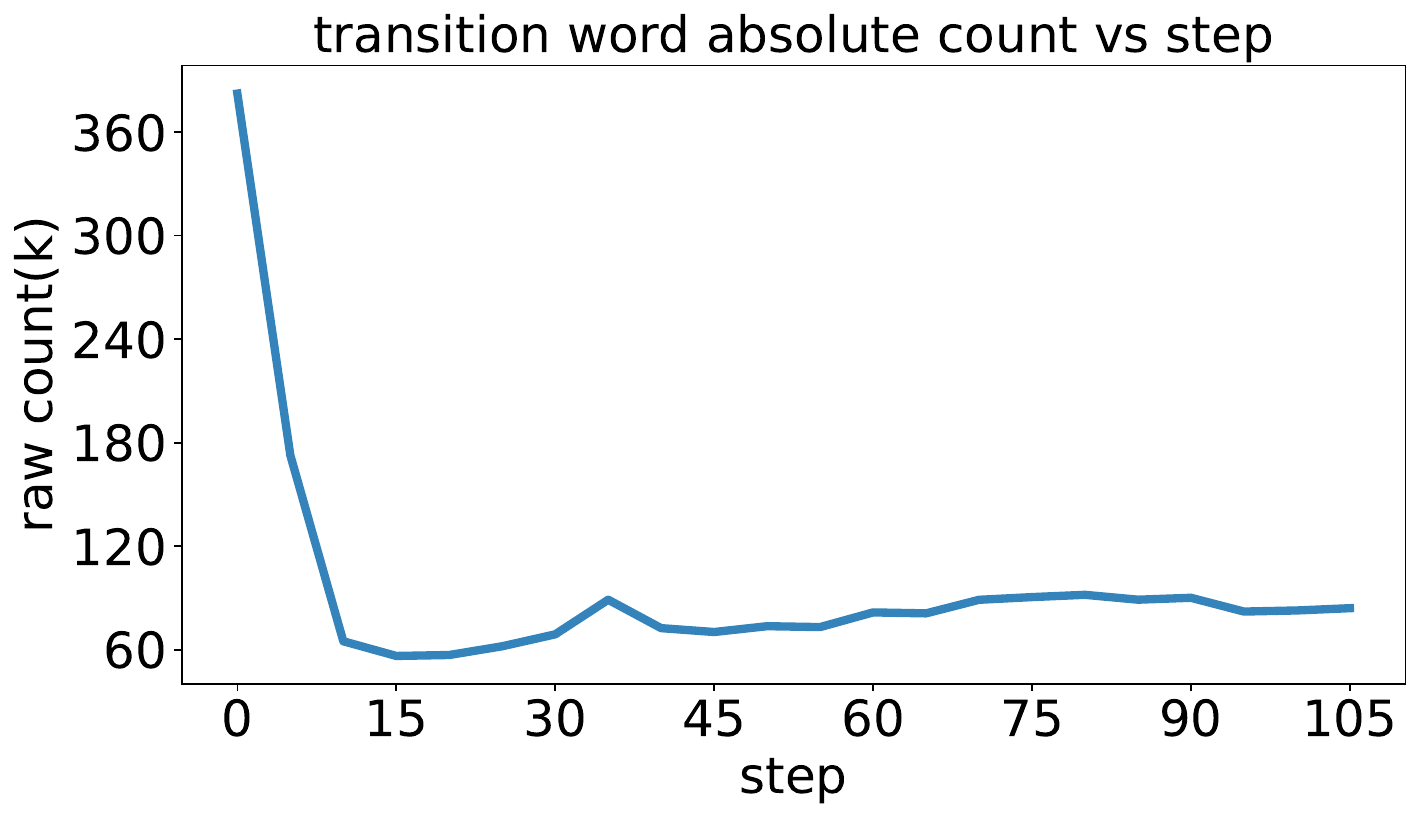} \hfill
  \includegraphics[width=0.5\linewidth]{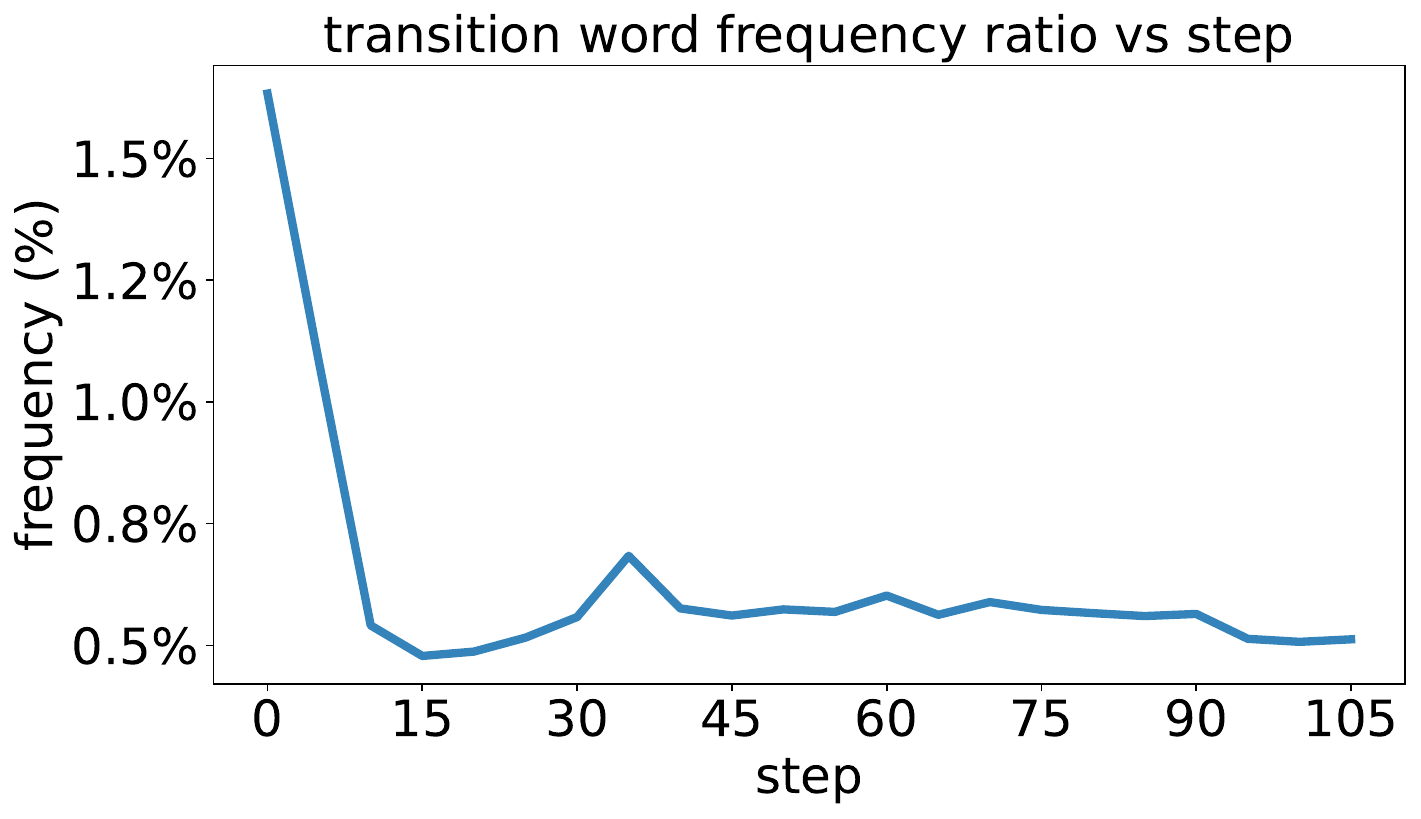}
  \caption {Counts(left) and Frequencies(right) of transition/backtracking markers (e.g., \textit{but}, \textit{however}, \textit{wait}) in sampled outputs across JudgeRLVR generating stage training steps.}
  \label{fig:transition_word}
\end{figure*}

\section{Results}
\label{sec:results}

\subsection{Main Results}
Table~\ref{tab:main_results_base_gen_seq} compares Base SFT, Vanilla RLVR that optimizes only final-answer correctness, and our proposed JudgeRLVR (sequential two-stage). The overall conclusion is that, compared with Vanilla RLVR, JudgeRLVR achieves higher or comparable accuracy while producing substantially shorter generations on most benchmarks, demonstrating a better quality--efficiency trade-off.

\paragraph{In-domain math: improving accuracy while markedly reducing redundant reasoning.}
On AIME24/25, HMMT\_feb\_2025, and BeyondAIME, JudgeRLVR consistently improves accuracy over Vanilla RLVR, while significantly reducing generation length. This matches the motivation of our divided stages: judging stage, via strict balanced sampling (equal numbers of correct and incorrect traces per problem), strengthens the model's ability to recognize error patterns and unproductive reasoning; then in generating stage, generating optimization can more reliably follow higher-confidence, lower-branching solution paths, reducing trial-and-error backtracking and verbose exploration. The larger gains on HMMT and BeyondAIME further suggest that the advantage of ``judge first, then generate'' is more pronounced when problems require longer chains of reasoning and stronger strategy selection.

On MATH500, JudgeRLVR is slightly worse in accuracy than Vanilla RLVR, but reduces generation length drastically. This indicates that for relatively saturated datasets, there is limited headroom for further accuracy gains, while JudgeRLVR can still substantially reduce reasoning cost when accuracy is largely maintained.

\paragraph{Out-of-domain: both generalization and efficiency gains, with task-dependent trade-offs.}
On GPQA Diamond, LiveCodeBenchv6, and MMLU-Redux, JudgeRLVR improves accuracy while also reducing length. This suggests that the ``error-sensitive / avoid-detours'' preference learned in judging stage is not restricted to math, but can transfer to science reasoning, coding, and broad knowledge tasks, reducing ineffective exploration and improving overall decision quality.

In contrast, on IFEval and ZebraLogic we observe ``higher accuracy but longer outputs''. This suggests that for tasks emphasizing format compliance, constraint satisfaction, or discrete rule verification, the model may need to generate more explicit checks to ensure correctness, trading off some length advantage for a higher pass rates. This does not contradict our goal: JudgeRLVR is not designed to mechanically minimize length, but to reduce \emph{unproductive} reasoning; when a task intrinsically requires more explicit structure, a moderate increase in length may be a necessary cost.

\subsection{Ablation Results}
Table~\ref{tab:main_results_mixed_judgeonly_seq} reports two ablations---Judge Only and Mixed Strategy---to test: (\romannumeral1) whether judging stage alone is sufficient; and (\romannumeral2) whether sequential staging is essential.

\paragraph{Judge Only: better ``judging'' does not reliably translate into better generation.}
Judge Only shows clear degradations on multiple benchmarks, especially on in-domain math: compared with JudgeRLVR, accuracy drops on math benchmarks. More importantly, generation length increases substantially. This indicates that the discriminative capability learned in judging stage does not automatically induce a more concise generation policy; it can even make the model more prone to ``writing out the checking process'' or reasoning more cautiously and verbosely, which lengthens outputs and hurts final accuracy. In other words, error-aware discrimination is a necessary prerequisite, but not a substitute for generating optimization; generating stage is still required to convert error sensitivity into better path selection and more information-dense expression during generation.

We also observe that Judge Only can achieve comparable or slightly better accuracy on some out-of-domain tasks while producing shorter outputs. This suggests that some ``judging'' preferences learned in judging stage may directly benefit certain tasks; however, this benefit is unstable and cannot replace the full two-stage pipeline for our core objective.

\paragraph{Mixed Strategy: interleaving judge and generate updates weakens the benefit of sequential stages.}
Mixed Strategy can be close to JudgeRLVR on some tasks, but is overall more unstable: it slightly exceeds JudgeRLVR on AIME25 and MATH500, yet drops substantially on tasks such as IFEval and LiveCodeBenchv6, often with significantly longer outputs. This aligns with our hypothesis: when ``judge-style'' and ``generation-style'' behaviors are optimized in a 1:1 interleaved manner, the model must satisfy two different objectives and reward structures in the same phase, which can lead to interference in policy-gradient updates and prevent the emergence of a clean, stable internal decision process of ``judge first, then generate''. The result is occasional local accuracy gains, but weaker overall efficiency and generalization.

\subsection{Mechanism Verification}
\paragraph{Style transfer measured by PPL under Base SFT.}
Figure~\ref{fig:ppl_vs_step} reports the Base SFT model's average perplexity on sampled outputs across training steps. For Vanilla RLVR, the average PPL remains largely unchanged, suggesting the output distribution stays stylistically close to Base SFT. In contrast, during JudgeRLVR judging stage, Base SFT PPL increases noticeably, indicating a systematic style shift induced by judge training.

\paragraph{Reduced explicit backtracking measured by transition words.}
Figure~\ref{fig:transition_word} show that during JudgeRLVR generating stage, both the absolute count and frequency of transition/backtracking words decrease substantially over training, consistent with reduced explicit backtracking and a shift of verification/correction into internal decision-making.

\section{Conclusion}
In this paper, we proposed JudgeRLVR, a two-stage training paradigm that acts as a precursor to efficient reasoning. By explicitly training the model to discriminate solution validity before generating, we achieved a superior quality--efficiency trade-off. Our experiments on Qwen3-30B-A3B show that JudgeRLVR improves in-domain math accuracy while reducing generation length by approximately 42\%, with notable generalization to out-of-domain tasks.

Many assume that judging an answer is the easy part, because verification seems to come with a checklist that generation does not. Our study points to a sharper truth: on identical math problems with the same model, judging yields greater improvement than solving alone, so “is this correct?” is a problem in its own right. Training the judge makes the model internalize what a good line of thought looks like, and later generation can cross its own limits with fewer false starts. In this sense, to generate is simply to judge so well that the correct answer is the only thing left to say.

\bibliography{ref}
\appendix

\section{Prompt Templates of Judge and Generate}
\label{sec:prompt templates}

\textbf{Judging Prompt}

\begin{tcolorbox}[colback=gray!10,colframe=gray!50,boxrule=0.5pt,arc=2pt]
\texttt{Given the question:}\\
\texttt{\{question\}}\\[2pt]
\texttt{And the proposed answer:}\\
\texttt{\{solution response\}}\\[2pt]
\texttt{Analyze concisely whether the answer is correct. After your analysis, put only your final judgment in \textbackslash boxed\{\} as 1 for correct and 0 for incorrect.}
\end{tcolorbox}

\textbf{Generating and Rollout Prompt}

\begin{tcolorbox}[colback=gray!10,colframe=gray!50,boxrule=0.5pt,arc=2pt]
\texttt{\{question\}}\\[2pt]
\texttt{Let's think step by step and output the final answer within \textbackslash boxed\{\}.}
\end{tcolorbox}

\section{Case Study}
\label{sec:full_output}
Full output of Figure~\ref{fig:JudgeRLVR_Example_Case}

\paragraph{Problem Statement}
\begin{quote}
Convert the point $(0,3)$ in rectangular coordinates to polar coordinates.  Enter your answer in the form $(r,\theta),$ where $r > 0$ and $0 \le \theta < 2 \pi.$
\end{quote}

\paragraph{Vanilla RLVR}
\begin{quote}\small
<think>

I need to convert the point (0,3) from rectangular coordinates to polar coordinates. The rectangular coordinates are given as (x,y) = (0,3), and I need to find the polar coordinates (r, $\theta$), where r is the distance from the origin, and $\theta$ is the angle from the positive x-axis.

First, recall the formulas for conversion. The radius r is given by r = $\sqrt{(x^{2} + y^{2})}$. Plugging in the values, x is 0 and y is 3, so r = $\sqrt{(0^{2} + 3^{2})}$ = $\sqrt{(0 + 9)}$ = $\sqrt{9}$ = 3. So r is 3.

Now, for the angle $\theta$, it's given by $\theta = \tan^{-1}(y/x)$. But here, x is 0, and y is 3. So if I plug in, $\tan^{-1}(3/0)$, which is undefined because division by zero. That makes sense because when x is 0, the point is on the y-axis.

I know that when a point is on the positive y-axis, the angle $\theta$ should be $\pi/2$ radians, or 90 degrees. Since the problem asks for $\theta$ in radians between 0 and $2\pi$, I should use radians.

So, $\theta = \pi/2$.

Let me confirm. If r is 3, and $\theta$ is $\pi/2$, then the rectangular coordinates should be $x = r \cos \theta = 3 * \cos(\pi/2) = 3 * 0 = 0$, and $y = r \sin \theta = 3 * \sin(\pi/2) = 3 * 1 = 3$. Perfect, that matches.

The problem says r > 0, which is satisfied since r=3>0, and $\theta$ between 0 and $2\pi$, $\pi/2$ is within that range.

I recall that sometimes people might think of $\theta$ as $-\pi/2$ or $3\pi/2$, but that would give a negative y-value, which is not the case here. Since y is positive, it has to be in the first or second quadrant, but since x is 0, it's specifically on the positive y-axis, so $\theta=\pi/2$.

If $\theta$ were $3\pi/2$, that would be the negative y-axis, where y is negative, but here y is positive, so it's not that.

So, I think $(3, \pi/2)$ is the correct polar coordinate.

The answer should be in the form (r, $\theta$), so (3, $\pi/2$).

I should write it as a pair. And the problem says to box the final answer.

So, I think that's it.

Just to be thorough, is there any other representation? For example, could r be negative? But the problem specifies r > 0, so no. And $\theta$ is usually taken in $[0, 2\pi)$, so $\pi/2$ is fine.

Sometimes people use degrees, but here it's specified to use radians, and the range is given, so radians are implied.

The problem doesn't specify units, but in mathematics, especially with $\pi$, it's standard to use radians.

So, I'm confident.

</think>

First, I need to convert the point $(0, 3)$ from rectangular coordinates to polar coordinates $(r, \theta)$ with $r > 0$ and $0 \le \theta < 2\pi$, use the conversion formulas.

The radius $r$ is given by:
\[
r = \sqrt{x^2 + y^2} = \sqrt{0^2 + 3^2} = \sqrt{9} = 3.
\]

The angle $\theta$ is found using $\theta = \tan^{-1}(y/x)$. However, since $x = 0$ and $y = 3 > 0$, the point lies on the positive y-axis. Thus, $\theta = \pi/2$.

Verify the solution:
\begin{itemize}
  \item $x = r \cos \theta = 3 \cos(\pi/2) = 3 \cdot 0 = 0$,
  \item $y = r \sin \theta = 3 \sin(\pi/2) = 3 \cdot 1 = 3$,
\end{itemize}
which matches the original point.

The conditions are satisfied: $r = 3 > 0$ and $\theta = \pi/2$ is in $[0, 2\pi)$.

Thus, the polar coordinates are $(3, \pi/2)$.]

The final answer is:
\[
\boxed{\left(3,\ \dfrac{\pi}{2}\right)}
\]
\end{quote}

\paragraph{JudgeRLVR}
\begin{quote}\small
<think>

We need to convert (0,3) to polar coordinates. r = sqrt(x\^{}2 + y\^{}2) = sqrt(0+9)=3. Theta is angle from positive x-axis to point (0,3). That's 90 degrees = pi/2 rad. Since x=0, y>0, theta = $\pi/2$. So answer (3, $\pi/2$). Ensure within 0 to $2\pi$. So final answer \boxed{(3,\frac{\pi}{2})}.

</think>
\end{quote}

\begin{quote}\small
First, I need to calculate r.

The point $(0,3)$ has

\[
r = \sqrt{0^{2}+3^{2}} = 3
\]

and lies on the positive $y$-axis, so the angle from the positive $x$-axis is

\[
\theta = \frac{\pi}{2}.
\]

Thus the polar coordinates are $(3,\frac{\pi}{2})$.

The final answer is:
\[
\boxed{(3,\frac{\pi}{2})}
\]
\end{quote}

\section{Evaluate Details}
\label{sec:Evaluate Details}
Table~\ref{tab:benchmarks} shows the metric setup of evaluation benchmarks.

\begin{table}[t]
\centering
\caption{Evaluation Benchmarks}
\label{tab:benchmarks}
\begin{tabular}{lll}
\hline
\textbf{Benchmark} & \textbf{Metric} \\
\hline
AIME24 & Acc~avg@10 \\
AIME25 & Acc~avg@10 \\
MATH500 & Acc \\
HMMT\_feb\_2025 & Acc~avg@10 \\
BeyondAIME & Acc~avg@8 \\
GPQA Diamond & Exact Match~avg@5 \\
IFEval & Prompt Strict \\
LiveCodeBenchv6 & Pass@1 \\
MMLU-Redux & Exact Match \\
ZebraLogic & Acc \\
\hline
\end{tabular}
\end{table}

\end{document}